\documentclass[10pt,twocolumn,letterpaper]{article}

\usepackage{iccv}              

\definecolor{iccvblue}{rgb}{0.21,0.49,0.74}
\usepackage[pagebackref,breaklinks,colorlinks,allcolors=iccvblue]{hyperref}

\usepackage{hyperref}       
\usepackage{url}            
\usepackage{booktabs}       
\usepackage{amsfonts}       
\usepackage{nicefrac}       
\usepackage{microtype}      
\usepackage{xcolor}         
\usepackage{amsmath}
\usepackage{amssymb}
\usepackage{algpseudocode}
\usepackage{fp}
\usepackage{lipsum,booktabs,epstopdf}
\usepackage{lipsum}
\usepackage{rotating}
\usepackage{multirow}
\usepackage{bbold}
\usepackage{algpseudocode}
\usepackage{pifont}
\newcommand{\cmark}{\ding{51}}
\newcommand{\xmark}{\ding{55}}
\usepackage{array, makecell} 
\usepackage{xcolor,colortbl}
\usepackage{comment}
\usepackage{kantlipsum}
\usepackage{tabularx}
\usepackage{booktabs}       
\usepackage{amsmath}
\usepackage{diagbox}
\usepackage{amsfonts} 
\usepackage{caption}
\usepackage{etoolbox}
\usepackage{enumitem}

\usepackage{graphicx}
\usepackage{subcaption}
\usepackage{float}
\usepackage{soul}
\usepackage[accsupp]{axessibility}  

\def\paperID{886} 
\def\confName{ICCV}
\def\confYear{2025}

\title{Visual Intention Grounding for Egocentric Assistants}

\author{
Pengzhan Sun\textsuperscript{1} \hspace{3mm}
Junbin Xiao\textsuperscript{1} \textsuperscript{*} \hspace{3mm}
Tze Ho Elden Tse\textsuperscript{1} \hspace{3mm}
Yicong Li\textsuperscript{1}\hspace{3mm}\\
Arjun Akula\textsuperscript{2}\hspace{3mm}
Angela Yao\textsuperscript{1} \hspace{3mm}\\%
\textsuperscript{1}National University of Singapore \hspace{3mm}
\textsuperscript{2}Google DeepMind\hspace{3mm} \\
\tt\small \{pengzhan, junbin, eldentse, ayao\}@comp.nus.edu.sg, \\
\tt\small liyicong@u.nus.edu,
\tt\small arjunakula@google.com
}

\begin{document}


\twocolumn[{%
\vspace{-22pt}
\renewcommand\twocolumn[1][]{#1}%
\maketitle
\includegraphics[width=1.0\linewidth]{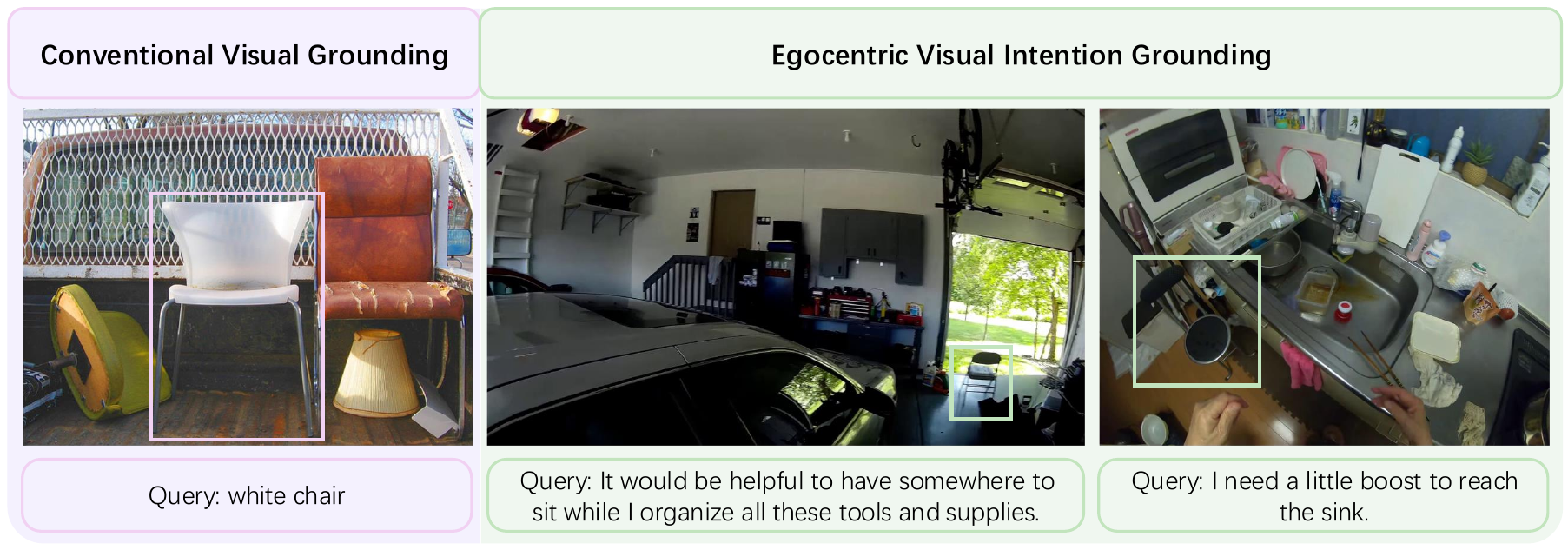}
\captionof{figure}{Traditional visual grounding (left) vs. egocentric visual intention understanding (center and right). Traditional grounding identifies the ``white chair" by detecting specific objects from third-person perspectives. Egocentric visual intention understanding must infer user needs in complex, first-person scenarios, \eg, 
seating in a workshop (center) or 
using a chair to reach the sink (right).}
\label{fig.task}
\vspace{16pt}
}]

\renewcommand{\thefootnote}{}
\footnotetext{\textsuperscript{*}Corresponding author}
\addtocounter{footnote}{-1}

\begin{abstract}
Visual grounding associates textual descriptions with objects in an image.
Conventional methods target third-person image inputs and named object queries. 
In applications such as AI assistants, the perspective shifts -- inputs are egocentric, and objects may be referred to implicitly through needs and intentions.
To bridge this gap, we introduce EgoIntention, the first dataset for egocentric visual intention grounding.
EgoIntention challenges multimodal LLMs to 1) understand and ignore unintended contextual objects and 2) reason about uncommon object functionalities.
Benchmark results show that current models misidentify context objects and lack affordance understanding in egocentric views. 
We also propose Reason-to-Ground (RoG) instruction tuning; it enables hybrid training with normal descriptions and egocentric intentions with a chained intention reasoning and object grounding mechanism. 
RoG significantly outperforms naive finetuning and hybrid training on EgoIntention,
while maintaining or slightly improving naive description grounding.
This advancement enables unified visual grounding for egocentric and exocentric visual inputs while handling explicit object queries and implicit human intentions.
Our code and model are available at \href{https://github.com/pengzhansun/EgoIntention}{https://github.com/pengzhansun/EgoIntention}.

\end{abstract}

\section{Introduction}
\label{sec:intro}

Consider the following scenarios: a person is looking for a place to sit down for organizing tools in a messy workshop, or a kid is trying to reach the sink in a kitchen (refer to Figure~\ref{fig.task}). A wearable artificial intelligent (AI) assistant could enhance these tasks by identifying contextually relevant objects (\eg, a chair) without explicit object references. 
To achieve this, such an assistant must possess strong egocentric visual perception capabilities~\cite{grauman2022ego4d, zhou2025egotextvqa, xiao2025egoblind, sun2021counterfactual}. This would significantly improve task efficiency, reduce cognitive load, and support hands-free, context-aware interaction in dynamic environments.

\begin{figure}[t]
\centering
\includegraphics[width=0.9\linewidth]{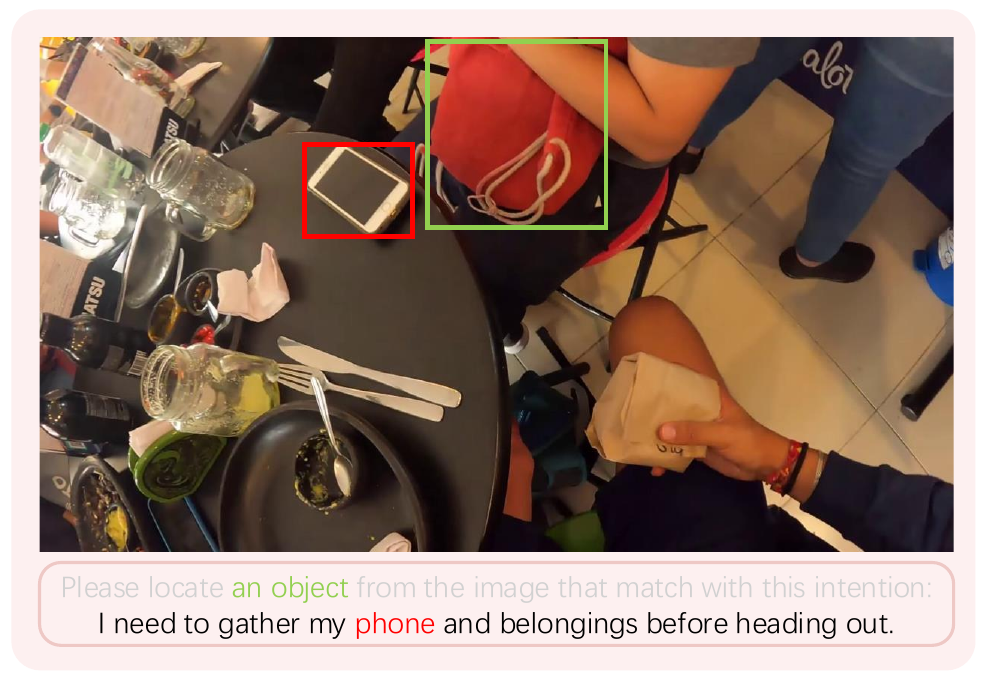}
\caption{Challenge of Visual Intention Grounding. The model must infer the intended object from the full intention sentence, rather than simply detecting explicitly mentioned objects. In this example, ``gather my phone and belongings'' explicitly mentions ``phone'' (highlighted in red) , which often misleads existing visual grounding models to identify the wrong object (red box). The correct target, a handbag (green box), is only implied. 
}
\label{fig.challenge}
\vspace{-0.2cm}
\end{figure}

Building on this vision, we introduce the egocentric visual intention grounding task.
Given an egocentric visual input and a human intention query, a model must accurately localize the intended object within the scene.
This task supports real-world applications where users locate objects based on their needs.
The object may not be directly named but can be inferred from the user’s intention. Additionally, egocentric AI assistants operate from a first-person perspective, introducing challenges such as occlusions and dynamic viewpoints not present in conventional third-person vision systems.
Unlike conventional visual grounding tasks~\cite{yu2016modeling,mao2016generation,nagaraja2016modeling}, such as referring expression comprehension~\cite{van2006building,viethen2008use,golland2010game,mitchell2010natural,mitchell2013generating,fitzgerald2013learning,kazemzadeh2014referitgame}, this task requires reasoning about object affordance beyond explicit mentions of the object. As shown in Figure~\ref{fig.challenge}, existing models often misinterpret explicit mentions and fail to infer the actual intended object, highlighting the need for contextual reasoning in visual intention grounding.

Despite recent advancements in vision-language models~\cite{chen2023shikra,you2023ferret,zhang2023next,zhang2024llava,li2024groundinggpt,shao2024visual,pramanick2024jack,chen2024lion,shi2024math,fan2025chestx,li2022invariant}, existing methods~\cite{chen2023minigpt,bai2023qwen,wang2024qwen2,achiam2023gpt,wang2023cogvlm,ma2025groma} struggle to solve visual intention grounding in diverse real-world scenarios.
First, solving the visual intention grounding task using two separate models, such as a language model (\eg, GPT-4~\cite{achiam2023gpt}) for reasoning and an object detector (\eg, GroundingDINO~\cite{liu2023grounding}) for localization, yields suboptimal results.
These off-the-shelf models process visual and language information independently, often resulting in inconsistent associations between the two modalities. Consequently, a model may hallucinate objects by incorrectly identifying items that are not present in the scene.
Secondly, while multimodal large language models (MLLMs) offer a unified approach, existing methods are primarily designed for third-person visual grounding tasks.
Without datasets connecting egocentric visual data to intention sentences, these models struggle to adapt to and perform well on first-person intention grounding.

To bridge the gap in egocentric visual intention grounding, we introduce EgoIntention. EgoIntention is a comprehensive dataset built upon Ego4D~\cite{grauman2022ego4d}, the largest real-world egocentric vision dataset.
We inherit object bounding boxes annotation from PACO-Ego4D~\cite{ramanathan2023paco}, a dataset that annotates object parts and attributes, and augment them with carefully curated human intention descriptions. Additionally, we address the inherent subjectivity in intention-object relationships by incorporating supplementary bounding box annotations for alternative objects that could reasonably fulfill each stated intention.
Our EgoIntention dataset addresses two main gaps in visual grounding research: 1) the lack of egocentric data, and 2) the limited coverage of complex intentions arising from first-person viewpoints.
Comprising 26,384 images, 52,768 human intention descriptions, and 89,841 annotated target object bounding boxes, this comprehensive dataset establishes a robust foundation for advancing visual intention understanding in real-world egocentric applications.

Despite providing a rich benchmark for egocentric intention grounding, EgoIntention also exposes significant challenges for existing models. 
We identify two major limitations.
First, multimodal models rely on explicit object-centric prompts (\eg, ``Where is the white chair?"), directly mapping object names to their locations. However, intention-based queries require implicit reasoning. For example, when assisting in a workshop, the model must infer that the user needs a chair for sitting, rather than simply detecting a chair.
Second, models must reason beyond direct object matches, recognizing alternative objects based on affordances. In the case of a child trying to reach the sink, the model should identify the chair as a suitable support, even though the query does not explicitly mention ``chair." This requires a deeper understanding of object functionality and context, which is a challenge for current MLLMs.

To address the above challenges, we propose a model-agnostic Reason-to-Ground (RoG) instruction tuning approach.  RoG disentangles intention understanding from object grounding to enhance multimodal models' reasoning capabilities. By doing so, RoG reduces spurious correlations between unintended object mentions and the actual intended object locations.
On EgoIntention, our RoG instruction tuning improves MiniGPTv2’s performance by 3.9 Precision@0.5 compared to naive finetuning and significantly outperforms the off-the-shelf GPT-4~\cite{achiam2023gpt} + GroundingDINO pipeline by 12.2 Precision@0.5.

Our contributions can be summarized as follows:
\begin{enumerate}

\item We construct the EgoIntention dataset for visual intention grounding. 
The dataset is the first egocentric visual grounding dataset with multiple intention queries.

\item We benchmark and reveal that existing MLLMs struggle with intention reasoning and egocentric visual grounding. 
These models misinterpret explicitly mentioned objects as targets or hallucinate objects not present in the scene.

\item We propose Reason-to-Grounding (RoG) instruction tuning, a model-agnostic training approach to enhance MLLMs for egocentric intention grounding while retaining their performances for normal visual grounding.
\end{enumerate}




\section{Related Work}

\begin{table}[t!]
\centering
\caption{Comparison of intention-related visual grounding datasets.}
\scalebox{0.75}{
\begin{tabular}{cccccccc}
    \toprule[1.5pt]
    Dataset & \#Images & \makecell[l]{Language \\ Query} & Ego-view & \makecell[l]{Multi-intention \\ annotations} \\
    \midrule[0.75pt]
    \textbf{ADE-Aff}~\cite{chuang2018learning} & 10,000 & Verb & \xmark & \xmark  \\
    \textbf{PAD}~\cite{luo2021one} & 4,002 & Verb & \xmark & \xmark \\
    \textbf{COCO-Tasks}~\cite{sawatzky2019object} & 39,724 & Phrase & \xmark & \xmark \\
    \midrule[0.75pt]
    \textbf{RIO}~\cite{qu2024rio} & 40,214 & \makecell[l]{Template \\ language} & \xmark & \cmark \\
    \midrule[0.75pt]
    \textbf{IntentionVG}~\cite{wang2024beyond} & 101,648 & \makecell[l]{Free-form \\ language} & \cmark & \xmark \\
    \midrule[0.75pt]
    \textbf{\makecell[l]{EgoIntention}} & 26,384 & \makecell[l]{Free-form \\ language} & \cmark & \cmark \\
  \bottomrule[1.5pt]
\end{tabular}
}
\label{tab_bias_metric_comparision}
\end{table}

\subsection{Visual Grounding Datasets}





Visual grounding~\cite{yu2016modeling,mao2016generation,nagaraja2016modeling} is a multimodal task that locates a target object in an image based on a given language query.
Early works 
focused on referring expression comprehension~\cite{van2006building,viethen2008use,golland2010game,mitchell2010natural,mitchell2013generating,fitzgerald2013learning,kazemzadeh2014referitgame}, which matches descriptive phrases to objects within an image. Datasets such as RefCOCO~\cite{kazemzadeh2014referitgame}, RefCOCO+~\cite{yu2016modeling}, and RefCOCOg~\cite{mao2016generation} have played key roles in advancing this field.
More recently, the scope of language input has been widened to 
encompass descriptions of object affordance~\cite{li2024laso,bahl2023affordances,yoshida2024text}. This evolution led to datasets using verbs or phrases, (\eg, ADE-Aff~\cite{chuang2018learning}, PAD~\cite{luo2021one}, COCO-Tasks~\cite{sawatzky2019object}) and full sentences in datasets like RIO~\cite{qu2024rio} and IntentionVG~\cite{wang2024beyond}.

Compared to IntentionVG, where egocentric images are captured with a fixed viewpoint centered on objects, our dataset is constructed from Ego4D~\cite{grauman2022ego4d}, introducing greater visual challenges such as motion blur, small object sizes, and perspective distortions inherent to first-person vision.
For language queries, our dataset provides multiple intention sentences per object, reflecting the diverse ways an object can be used to fulfill different needs. Similar to the RIO dataset, we annotate each sample with both a context sentence, describing an object's typical use in its expected environment, and an uncommon sentence, which represents a less conventional use case requiring creative object substitution.


%



\subsection{Visual Grounding Models}


Traditional visual grounding methods~\cite{mao2016generation,yang2019dynamic,lin2014microsoft,liao2020real,luo2020multi} are 
specialized models explicitly trained to map language queries to object locations. Models such as MDETR~\cite{kamath2021mdetr}, SeqTR~\cite{zhu2022seqtr}, and Polyformer~\cite{liu2023polyformer} leverage Transformer-based architectures to enhance this association.
GroundingDINO~\cite{liu2023grounding}, a recent advancement in open-set object detection, extends DINO~\cite{zhang2022dino} with grounded pre-training, allowing it to detect arbitrary objects given category names or referring expressions.
Recently, multimodal large language models (MLLMs) have emerged as the dominant paradigm for vision-language tasks~\cite{chen2023minigpt,bai2023qwen,wang2024qwen2,achiam2023gpt,wang2023cogvlm,ma2025groma, butd}. By leveraging vast amounts of image-text data and instruction tuning, these models achieve impressive generalization across various multimodal benchmarks.
While effective in conventional visual grounding tasks, both specialist models and MLLMs primarily perform word-by-word detection on the input language query, rather than truly understanding the underlying human intention. 
To address these challenges, we propose Reason-to-Ground (RoG), a method that explicitly disentangles intention reasoning from object localization, as detailed in method Section~\ref{sec:method}.

\section{Intention Grounding \& EgoIntention Dataset}

\begin{figure*}[h]
\centering
\includegraphics[width=1.0\linewidth]{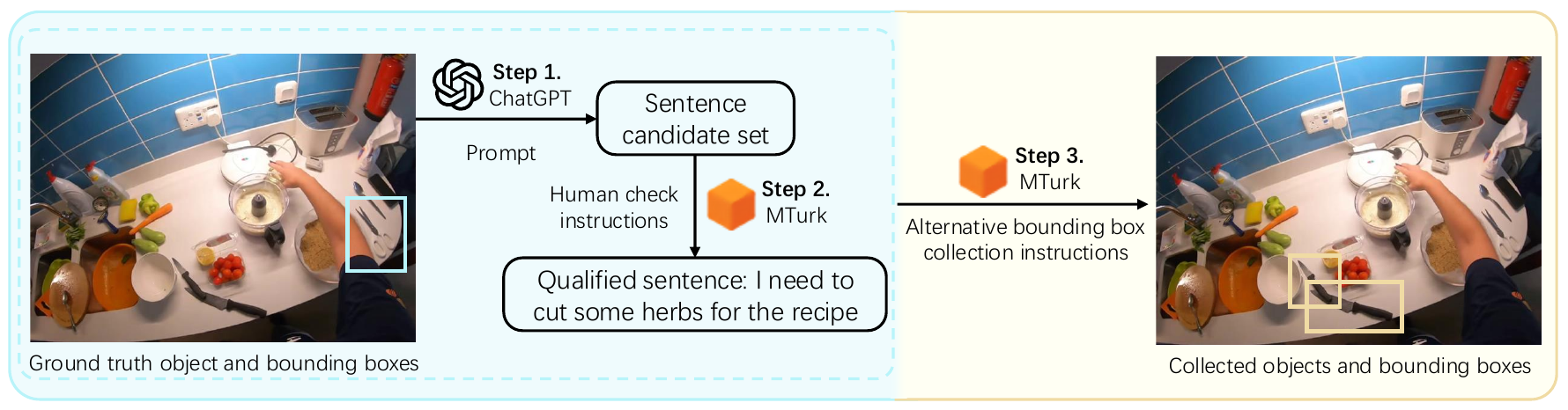}
\caption{Dataset collection pipeline for EgoIntention. Our data collection process consists of three key stages. (1) Intention Sentence Generation: We use GPT-4 to generate egocentric human intention sentences based on visual input, covering both context-aware and uncommon intention scenarios. (2) Human Validation via MTurk: Annotators assess the semantic validity and real-world applicability of generated sentences, filtering out low-quality or ambiguous descriptions. (3) Alternative Object and Bounding Box Collection: Given the inherent subjectivity of human intentions, additional valid object candidates are identified by human annotators, supplementing the original ground truth annotations with alternative bounding boxes.
}
\label{fig.dataset}
\end{figure*}

\subsection{Task Description}

We introduce visual intention grounding, a novel paradigm that establishes a direct mapping between human intentions and target objects in visual scenes. This approach bridges the gap between natural language understanding and visual object grounding.
Formally, given an egocentric image $I$ and a human intention query $Q$, the task requires the model to comprehend the underlying intention and localize the target object $O$ that satisfies the user's need by predicting its bounding box coordinates $(x_1,y_1,x_2,y_2)$.

\subsection{Dataset Collection}

Our dataset, EgoIntention, sources its images from the Ego4D dataset~\cite{grauman2022ego4d} and its associated object bounding box annotations from PACO~\cite{ramanathan2023paco}.
The key contribution of EgoIntention is the collection of multiple intention queries per object, reflecting the diverse ways objects are used in real-world scenarios.
Our systematic data collection pipeline consists of three stages, as illustrated in Figure~\ref{fig.dataset}.

The initial stage generates intention descriptions. We use GPT-4~\cite{achiam2023gpt}'s multimodal capabilities to analyze egocentric visual inputs and generate contextually relevant human intention sentences. 
To capture diverse real-world scenarios, we design two types of intention descriptions:
\begin{itemize}
    \item \textbf{Context-aware intentions:} These sentences reflect complex object relationships and environmental cues from a first-person perspective. For instance, a user might think, ``I noticed a wobbly table leg that needs fixing,'' expressing the need for a hammer.  
    \item \textbf{Uncommon intentions:} These describe atypical object uses, where users repurpose objects based on necessity. For example, a backpack might be used as an improvised umbrella during unexpected rain.  
\end{itemize}
These sentences are generated using carefully crafted prompts through an in-context learning approach as shown in Appendix.

The second stage involves human validation via Amazon Mechanical Turk (MTurk). Annotators assess each generated sentence using detailed guidelines to ensure semantic correctness and real-world plausibility. Since multiple valid intentions may exist for a given scenario, we do not use inter-annotator agreement as a filtering criterion. Instead, one annotator selects and, if necessary, refines a suitable sentence. This selection is then verified by a second annotator and a GPT-4-based checker, both of which independently assess whether the sentence expresses a genuine need for the target object. Only those passing both checks are retained.

To account for the subjectivity of human intentions, we include additional object annotations. A single intention may be satisfied by various objects (\eg, flower pots, bottles, or cups for desktop decoration). Annotators identify such alternatives, which are added as supplementary bounding boxes. As before, only those passing both human and GPT-4 verification are included.

We observed that GPT-4 performs more reliably as a verifier than as a generator. While context-aware sentences passed human checks at 97.2\%, uncommon intentions had a lower pass rate of 74.1\%, often due to generic outputs or object name leakage. However, as a verifier, GPT-4 achieved 92\% agreement with human judgments in a 500-sample study, supporting its role as an auxiliary checker. To further validate our pipeline, we conducted an inter-annotator study on 500 verified samples, yielding agreement rates of 98.6\% for context-aware and 91.2\% for uncommon intentions. The maintained or slightly improved RefCOCO performance after joint training further supports the quality of our annotations.


\subsection{Dataset Statistics}


\begin{table}[ht]
    \centering
    \begin{tabular}{lccc}
        \toprule
        & Image & Context BBox & Uncommon BBox \\
        \midrule
        Train & 15,667 & 25,772 & 25,933 \\
        Val   & 825 & 1,402 & 1,366 \\
        Test  & 9,892 & 17,699 & 17,669 \\
        \bottomrule
    \end{tabular}
    \caption{Number of images and bounding boxes (BBox) in the EgoIntention dataset. 
Context BBox refers to bounding boxes associated with objects commonly used in the given scene, aligning with expected human intentions based on environmental cues. 
Uncommon BBox represents bounding boxes for objects used in unconventional ways, where the intended action requires creative or atypical object usage.}
    \label{tab:dataset_summary}
\end{table}






Our EgoIntention dataset builds upon PACO's image splits, comprising 15,667 training, 825 validation, and 9,892 test samples. For each image, we annotate two types of intention queries: context-aware intentions that leverage environmental cues, and uncommon intentions that capture alternative object uses. Following our multi-stage annotation pipeline, we enrich the dataset with supplementary bounding box annotations to accommodate the inherent diversity of object choices for each intention. The distribution of bounding box annotations across different splits and intention types is summarized in Table~\ref{tab:dataset_summary}.


\section{Method}

\label{sec:method}







This section begins with an observational study of off-the-shelf hybrid models in Section~\ref{sec:4.1}, highlighting key limitations in their reasoning and detection pipeline. Based on these findings, we propose Reason-to-Ground (RoG) to enhance multimodal large language models for visual intention grounding in Section~\ref{sec:4.2}. We then detail our supervised fine-tuning approach in Section~\ref{sec:4.3}.

\subsection{Observation of Off-the-Shelf Hybrid Models}

\label{sec:4.1}


A straightforward approach to visual intention grounding is to leverage two off-the-shelf models separately for reasoning and detection.
Specifically, we use ChatGPT for intention reasoning and GroundingDINO for object detection.
From our observation study (detailed in Section~\ref{sec:exp}), we identify two key findings:
Performing reasoning first improves accuracy. Narrowing the search space before detection leads to more precise object localization.
Hybrid off-the-shelf models suffer from inconsistent representations. These models operate in distinct visual and language spaces, causing discrepancies that hinder effective grounding.
Motivated by these findings, our main method introduces Reason-to-Ground, an instruction tuning approach for multimodal large language models.



\subsection{Reason-to-Ground Instruction Tuning}

\label{sec:4.2}

\begin{figure}[h]
\centering
\includegraphics[width=0.9\linewidth]{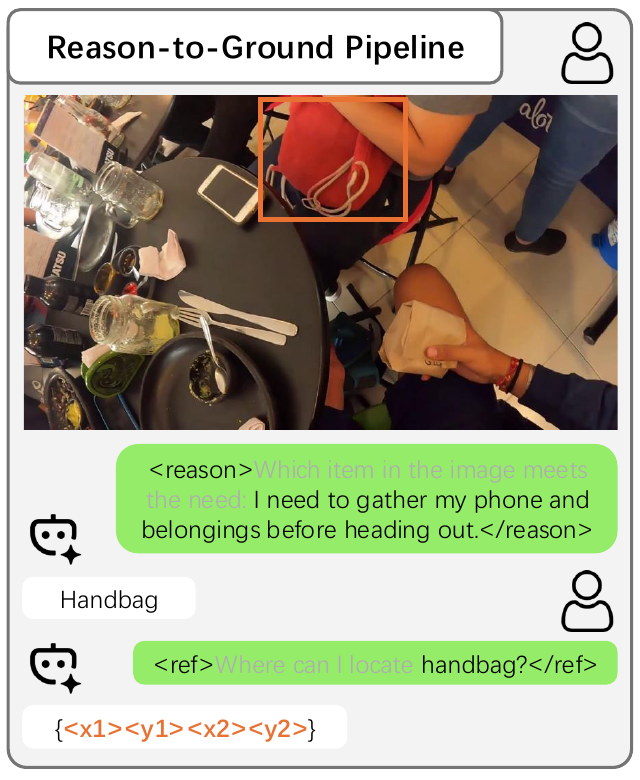}
\caption{Overview of Reason-to-Ground Instruction Tuning (RoG): The model first infers an explicit object category from an implicit intention sentence (intention reasoning), then localizes the object in the image (object grounding). 
}
\label{fig.method}
\end{figure}

We propose a novel training strategy, Reason-to-Ground Instruction Tuning (RoG).
Our method decomposes the task into two essential components: human intention understanding and visual grounding.
Existing approaches~\cite{chen2023minigpt,bai2023qwen} directly feed implicit intention sentences with a task-specific token \verb|<ref>|, which prompts the MLLM to output the bounding box corresponding to the input language query.
In contrast, RoG employs a two-stage process. In the first stage, we facilitate human intention understanding by querying the MLLM with a \verb|<reason>| token followed by the implicit intention sentence. This prompts the model to output the target object category.
In the second stage, we perform \verb|<ref>| token and the explicit object description derived from the first stage. These two stages are illustrated with an example in Figure~\ref{fig.method}.

By disentangling \textbf{intention reasoning} and \textbf{object grounding}, RoG prevents the model from naively associating explicitly mentioned objects with target bounding boxes, leading to more accurate intention-driven visual grounding.

\subsection{Supervised Fine-tuning}

\label{sec:4.3}

We propose a unified 
grounding framework that can process exo- and egocentric visual inputs while handling explicit object queries and implicit human intentions. To achieve this comprehensive capability, we leverage traditional exocentric datasets (RefCOCO~\cite{kazemzadeh2014referitgame}, RefCOCOg~\cite{mao2016generation}, and RefCOCO+~\cite{yu2016modeling}) alongside our proposed EgoIntention dataset during training.
Our method RoG effectively extends MLLMs' visual grounding capabilities to accommodate egocentric image inputs and implicit human intention queries through LoRA~\cite{hu2022lora} supervised fine-tuning.
\section{Experiments}

\label{sec:exp}

\subsection{Implementation Details}

We conducted supervised finetuning using a comprehensive dataset that combines traditional referring expression comprehension datasets (RefCOCO, RefCOCO+, and RefCOCOg) with our proposed visual intention grounding dataset, EgoIntention. For our experiments, we evaluated five state-of-the-art multimodal large language models (MLLMs), all with 7B parameters: MiniGPTv2, Groma, CogVLM-grounding, and Qwen-VL.
All model training was performed on 4 NVIDIA A40 GPUs. To maintain computational efficiency while preserving model performance, we employed Low-Rank Adaptation (LoRA)~\cite{hu2022lora} for parameter-efficient finetuning.

\subsection{Models}
\subsubsection{Hybrid Model Baselines}

A straightforward approach to visual intention grounding is to use off-the-shelf state-of-the-art models to separately address the vision and language components of the task. For vision, we adopt GroundingDINO, a popular open-set object detector that accepts a set of object category queries and returns the corresponding bounding boxes. For language understanding, we use ChatGPT, a large-scale language model developed by OpenAI, to reason about the user’s intention and infer the target object category.

As detailed in Section~\ref{sec:method}, we implement this two-stage framework in two variants: (1) first GPT-4 then GroundingDINO (Reasoning–Detection baseline, R-D), and (2) first GroundingDINO then GPT-4 (Detection–Reasoning baseline, D-R).

\textbf{D-R baseline.}
We begin by applying GroundingDINO to detect a comprehensive set of candidate object bounding boxes in the egocentric image. To ensure broad coverage, we use all object categories provided by the PACO dataset as text queries. GroundingDINO returns bounding boxes along with associated logits and predicted phrases. We then pass these predicted phrases, along with the human intention sentence, to GPT-4. GPT-4 performs reasoning over the textual descriptions to identify the object category most aligned with the expressed intention.

\textbf{R-D baseline.}
In contrast, this variant begins with GPT-4. Given the egocentric image and a prompt listing all candidate object categories, GPT-4 reasons about the intention and outputs the predicted object category. This output is then used to filter the query space for GroundingDINO, which subsequently predicts the bounding box for the inferred object by focusing only on the categories identified through GPT-4’s reasoning.

\subsubsection{Multimodal Large Language Models}
We evaluate our approach on several state-of-the-art MLLMs, categorized into two groups based on their specialized capabilities:

\textbf{Grounding Specialist MLLMs} excel at visual grounding tasks including grounded captioning, referring expression generation, referring expression comprehension, and grounded visual question answering:
\begin{itemize}
    \item \textbf{CogVLM-grounding} incorporates a trainable visual expert module within the attention and FFN layers, effectively bridging the gap between the frozen pretrained language model and image encoder.
    \item \textbf{Groma} demonstrates exceptional region-level grounding capability by integrating region tokens into both user instructions and model responses, enabling precise localization of described language queries.
\end{itemize}

Compared to grounding specialists, \textbf{Generalist MLLMs} exhibit enhanced reasoning abilities—a critical capability for solving visual intention grounding tasks:
\begin{itemize}
    \item \textbf{MiniGPT-v2} provides a unified interface for numerous vision-language tasks, employing unique task identifiers during model training to facilitate multi-task learning.
    \item \textbf{Qwen-VL}, built upon the Qwen-LM foundation, transcends conventional image description and question-answering capabilities by incorporating robust visual grounding functionality.
\end{itemize}

\subsection{Evaluation Metrics}

Our evaluation accounts for multiple alternative ground-truth boxes. 
We calculate each sample's score by computing the IoU between the predicted box and all ground truth boxes and then take the maximum IoU. A prediction is correct if its IoU with any ground truth box exceeds a threshold of 0.3 or 0.5 (Precision@0.3 or @0.5).

\subsection{Zero-shot Evaluation}

\begin{table*}[ht]
\centering
\caption{Benchmark comparison on EgoIntention across various methods, including two-stage pipeline approaches, grounding-specialist MLLMs, and generalist MLLMs. Performance is evaluated using Precision@0.3, Precision@0.5, and mIoU, reported for both the Context Split and Uncommon Split. The Overall P@0.5 metric summarizes the general performance across splits.}
\scalebox{1.0}{
\begin{tabular}{l|ccc|ccc|c}
\toprule
\multirow{2}{*}{\textbf{Method}} & \multicolumn{3}{c|}{\textbf{Context Split}} & \multicolumn{3}{c|}{\textbf{Uncommon Split}} & \multirow{2}{*}{\textbf{Overall P@0.5}} \\
\cmidrule(lr){2-4} \cmidrule(lr){5-7}
& \textbf{P@0.3} & \textbf{P@0.5} & \textbf{mIoU} & \textbf{P@0.3} & \textbf{P@0.5} & \textbf{mIoU} &  \\
\midrule
D-R GroundingDINO-GPT4 & 30.1 & 21.1 & 0.217 & 20.6 & 14.6 & 0.150 & 17.8 \\
R-D GPT4-GroundingDINO & 54.3 & 46.6 & 0.402 & 31.7 & 23.6 & 0.242 & 35.1 \\
\midrule
CogVLM-grounding & 8.0 & 5.9 & 0.057 & 6.3 & 4.9 & 0.042 & 5.4 \\
Groma & 9.6 & 7.4 & 0.074 & 8.9 & 6.9 & 0.070 & 7.2 \\
\midrule
MiniGPT-v2 & 31.3 & 21.7 & 0.224 & 25.5 & 18.0 & 0.186 & 19.9 \\
Qwen-VL & 27.9 & 21.4 & 0.225 & 27.3 & 22.0 & 0.228 & 21.7 \\
\bottomrule
\end{tabular}
}
\label{tab:zeroshot}
\end{table*}

\begin{table*}[ht]
\centering
\caption{Comparison of instruction tuning methods on RefCOCO, RefCOCO+, RefCOCOg, and EgoIntention datasets. Naive SFT refers to LoRA-based supervised fine-tuning, while RoG SFT represents LoRA-based fine-tuning with our Reason-to-Ground Instruction Tuning (RoG) instruction tuning approach. We report Precision@0.5 as the evaluation metric across different validation and test splits.}
\scalebox{1.0}{
\begin{tabular}{lccccccccccccc}
\toprule
\textbf{Model} & \multicolumn{3}{c}{\textbf{RefCOCO}} & \multicolumn{3}{c}{\textbf{RefCOCO+}} & \multicolumn{2}{c}{\textbf{RefCOCOg}} & \multicolumn{3}{c}{\textbf{EgoIntention}} \\
\cmidrule(lr){2-4} \cmidrule(lr){5-7} \cmidrule(lr){8-9} \cmidrule(lr){10-12}
& val & testA & testB & val & testA & testB & val & test & context & uncommon & overall \\
\midrule
Zero-shot MiniGPTv2 & 87.4 & 91.3 & 83.7 & 79.0 & 85.1 & 72.8 & 83.5 & 84.1 & 21.7 & 18.0 & 19.9 \\
\rowcolor[gray]{0.9} 
Naive SFT & 86.6 & 91.0 & 83.0 & 79.0 & 84.9 & 72.0 & 82.6 & 84.2 & 46.0 & 40.9 & 43.4 \\
\rowcolor[gray]{0.9}
RoG SFT & 87.8 & 91.4 & 84.0 & 79.8 & 85.4 & 73.8 & 84.3 & 85.2 & \textbf{49.9} & \textbf{44.7} & \textbf{47.3} \\
\midrule
Zero-shot Qwen-VL & 89.3 & 92.4 & 85.4 & 83.2 & 88.2 & 77.2 & 85.3 & 85.6 & 21.4 & 22.0 & 21.7 \\
\rowcolor[gray]{0.9} 
Naive SFT & 89.5 & 92.8 & 85.7 & 83.4 & 88.8 & 77.8 & 85.9 & 86.3 & 32.1 & 26.1 & 29.1 \\
\rowcolor[gray]{0.9}
RoG SFT & 89.3 & 92.5 & 85.3 & 83.3 & 88.8 & 77.4 & 86.2 & 86.4 & \textbf{35.5} & \textbf{31.7} & \textbf{33.6} \\
\bottomrule
\end{tabular}
}
\label{tab:main}
\end{table*}

\begin{table*}[ht]
\centering
\small
\caption{Ablation study of training datasets used for MiniGPT-v2 fine-tuning and their impact on visual grounding datasets.}
\scalebox{0.9}{
\begin{tabular}{cccclcccccccccccc}
\toprule
\multicolumn{3}{c}{\textbf{SFT Datasets}} & \multirow{2}{*}{\textbf{Method}} & \multicolumn{3}{c}{\textbf{RefCOCO}} & \multicolumn{3}{c}{\textbf{RefCOCO+}} & \multicolumn{2}{c}{\textbf{RefCOCOg}} 
& \multicolumn{4}{c}{\textbf{EgoIntention}} \\
\cmidrule(lr){1-3} \cmidrule(lr){5-7} \cmidrule(lr){8-10} \cmidrule(lr){11-12} \cmidrule(lr){13-16}
\textbf{RC/+/g} & \textbf{RCInt./+/g} & \textbf{EgoInt.} & & Val & TestA & TestB & Val & TestA & TestB & Val & Test & Con & Unco & Ave. & \textcolor{lightgray}{Obj.} \\
\midrule
\multicolumn{3}{c}{-} & 0-shot & 87.4 & 91.3 & 83.7 & 79.0 & 85.1 & 72.8 & 83.5 & 84.1 & 21.7 & 18.0 & 19.9 & \textcolor{lightgray}{40.8} \\
\midrule
\checkmark &  &  & Naive SFT & 87.6 & 91.3 & 84.4 & \textbf{80.0} & 85.3 & \textbf{73.8} & \textbf{84.8} & 85.2 & 23.7 & 19.4 & 21.5 & \textcolor{lightgray}{38.1} \\
 &  & \checkmark & Naive SFT & 66.5 & 71.5 & 60.4 & 60.2 & 66.6 & 51.8 & 64.6 & 65.8 & 42.8 & 39.2 & 41.0 & \textcolor{lightgray}{46.2}\\
\checkmark &  & \checkmark & Naive SFT & 87.5 & \textbf{91.5} & \textbf{84.6} & 79.9 & \textbf{85.6} & 73.5 & 84.7 & \textbf{85.4} & 45.9 & 40.8 & 43.3 & \textcolor{lightgray}{48.6} \\
\checkmark & \checkmark & \checkmark & Naive SFT & 86.6 & 91.0 & 83.0 & 79.0 & 85.0 & 72.0 & 82.6 & 84.2 & 46.0 & 40.9 & 43.4 & \textcolor{lightgray}{51.3}\\
\checkmark & \checkmark & \checkmark & RoG SFT & \textbf{87.8} & 91.4 & 84.0 & 79.8 & 85.4 & 73.8 & 84.3 & 85.2 & 49.9 & 44.7 & 47.3 & \textcolor{lightgray}{52.2} \\
\bottomrule
\end{tabular}
}
\label{tab:ablation}
\end{table*}

We first test all the 6 above mentioned methods in a zero-shot setting in Table~\ref{tab:zeroshot}. Our experiments reveal that \textbf{the order of using the reasoner and detector significantly impacts results}, despite the D-R and R-D baselines using the same models. For visual context-aware intention understanding, the D-R baseline correctly understands and accurately detects only 21.1\% of samples. In contrast, the R-D baseline improves P@0.5 accuracy to 46.6\%.
We attribute this difference to GroundingDINO's limitations when processing numerous language queries. When used first, it must handle all 
the object categories from the EgoIntention dataset and omits many object candidates. However, by using GPT-4 for initial reasoning, we narrow the target objects to one or two categories. This focused input allows GroundingDINO to perform 25\% better than in the D-R baseline.
For uncommon intention understanding, we observe a similar trend. The R-D baseline (23.6\%) outperforms the D-R baseline (14.6\%), but P@0.5 improves by only 9\%. This smaller improvement stems from GPT-4's lower reasoning accuracy for uncommon intentions, failing to provide GroundingDINO with the correct object category for detection.

\textbf{Grounding-specialist MLLMs, such as CogVLM-grounding and Groma, perform poorly on EgoIntentions.} This can be attributed to human intention reasoning being crucial for solving the visual intention grounding task. However, these specialist models are primarily aligned with grounding-related tasks and lack the necessary reasoning capability to infer which object should be grounded. Consequently, they fail to identify the intended object, leading to suboptimal results.
In contrast, \textbf{generalist MLLMs such as MiniGPT-v2 and Qwen-VL demonstrate more reasonable grounding performance} when given an intention query. 
Despite reasoning capabilities learned from pretraining and alignment, generalist MLLMs' overall performance remains limited. MiniGPT-v2 achieves an overall Precision@0.5 score of 19.9\%, while Qwen-VL reaches 21.7\%; both are significantly lower than the R-D GPT4-GroundingDINO baseline (35.1\%). This performance gap highlights the importance of integrating strong intention reasoning with visual grounding for improved results.

\subsection{RoG Supervised Finetuning Results}

In addition to the EgoIntention training set, we incorporate RefCOCO, RefCOCO+, and RefCOCOg (RefCOCO/+/g) to maintain model performance on referring expression comprehension while also leveraging these datasets to enhance intention grounding performance. As a result, our training data consists of three components: RefCOCO/+/g, RefCOCOIntention/+/g, and EgoIntention. The RefCOCOIntention/+/g dataset is generated automatically using GPT-4, applying the same prompt used for collecting EgoIntention, to create human intention queries.

While Naive SFT improves EgoIntention performance, it slightly degrades the model's capability in referring expression comprehension as shown in Table~\ref{tab:main}. In contrast, \textbf{our RoG instruction tuning strategy not only further enhances performance on EgoIntention but also leads to improved results on the RefCOCO series datasets.}
After fine-tuning with our RoG strategy, the generalist MLLM MiniGPT-v2 surpasses the best off-the-shelf two-stage method (R-D GPT4-GroundingDINO) by 12.2\% according to Precision@0.5, demonstrating the effectiveness of our approach in bridging intention reasoning and grounding.
Unlike R-D GPT4-GroundingDINO, which treats these as two independent sub-tasks, our method jointly models the reasoning and grounding processes, leading to more coherent and accurate results.
Additionally, we observe that RoG SFT performs better on MiniGPT-v2 than on Qwen-VL. This discrepancy arises because MiniGPT-v2 exhibits stronger adaptability to new task-specific tokens, enabling it to integrate hierarchical reasoning instructions more effectively. In contrast, Qwen-VL demonstrates weaker instruction-following capabilities during supervised fine-tuning, limiting its performance gains under the same strategy.
See visualization examples in Appendix.

\subsection{Ablation Study}

\subsubsection{Impact of Supervised Finetuning Datasets}

We analyze the performance gains from different datasets used during the SFT stage in Table~\ref{tab:ablation}. 
Finetuning exclusively on EgoIntention hurts generalization, dropping RefCOCO validation performance from 87.4\% to 66.5\%.
Combining RefCOCO/+/g with EgoIntention maintains REC performance while improving EgoIntention metrics. Specifically, context intention performance improves from 42.8\% to 45.9\%, and uncommon intention performance increases from 39.2\% to 40.8\% (see Table~\ref{tab:ablation}). This suggests that conventional REC datasets contribute additional gains, particularly in object grounding.

We explore whether adding human intention annotations to REC datasets improves EgoIntention performance. We create RefCOCOInt/+/g by collecting intention sentences for RefCOCO/+/g training sets. This augmentation yields only slight improvements. Without human verification, data quality remains a bottleneck. These limited gains suggest high-quality annotations are crucial for advancing intention grounding.

\textbf{The best overall performance comes from applying our RoG strategy across all training datasets}, demonstrating the effectiveness of our approach in jointly improving REC task and visual intention grounding task.

\subsubsection{RoG Improves Explicit Object Grounding}

As shown in Table~\ref{tab:ablation}, we further evaluate the model's performance on EgoIntention with explicit object queries, as indicated in the last column labeled ``object''. Compared to Naive SFT, RoG SFT disentangles visual intention understanding from object grounding, preventing the model from being misled by explicitly mentioned but unintended objects in the intention query. \textbf{Our approach also leads to a significant improvement in egocentric visual grounding with explicit object queries}, boosting the accuracy from 51.3\% with Naive SFT to 52.2\% with RoG SFT.

\subsubsection{Hallucination and Misleading Errors.}

Failure cases are due to \textit{hallucination} (object is neither mentioned in the query nor present in the image) and object reasoning. The latter can be divided into \textit{misleading language} (object mentioned in the query but not intended) and \textit{misleading vision} (object appears in the image but is not intended). 
We use RAM++~\cite{huang2023open} to extract all object tags in the image to check for misleading vision errors.  RoG fine-tuned MiniGPTv2 achieves a lower error rate across three categories as shown in the table~\ref{tab:error-analysis}.

\begin{table}[h]
  \centering
  \caption{Error rates (\%) for hallucination and reasoning failures across context and uncommon splits.}
  \small
  \scalebox{0.65}{
  \begin{tabular}{l|ccc|ccc}
    \toprule
    \multirow{2}{*}{Model} & \multicolumn{3}{c|}{Context Split} & \multicolumn{3}{c}{Uncommon Split} \\
    & \makecell[c]{Language \\ Mislead} & \makecell[c]{Vision \\ Mislead} & Hallucination & \makecell[c]{Language \\ Mislead} & \makecell[c]{Vision \\ Mislead} & Hallucination \\
    \midrule
    Zero-shot & 3.4 & 8.4 & 30.1 & 6.9 & 7.5 & 58.2 \\
    Naive SFT & \textbf{1.4} & 6.4 & 23.4 & 3.9 & 7.5 & 44.0 \\
    \rowcolor[gray]{0.9}
    RoG SFT & 2.1 & \textbf{3.8} & \textbf{12.0} & \textbf{2.4} & \textbf{6.6} & \textbf{24.3} \\
    \bottomrule
  \end{tabular}
  }
  \label{tab:error-analysis}
\end{table}

\subsubsection{Uncommon Intention Coverage}

Uncommon intentions are inherently subjective and underexplored. 
Prior work like RIO~\cite{qu2024rio} offers limited uncommon cases with 4,826 test samples only for testing and none for training.
We curated five diverse, uncommon samples per object category as in-context learning GPT-4 prompts to construct the first large-scale training set for this setting. Table~\ref{tab:uncommon-intention-results} shows that removing our uncommon split from training leads to a clear drop in Precision@0.5 on EgoIntention uncommon test set. Training with our uncommon intention data also improves zero-shot performance on RIO uncommon set for both naive and RoG SFT models.

\begin{table}[h]
  \centering
  \caption{Impact of training with uncommon intentions on EgoIntention and RIO benchmarks.}
  \small
  \scalebox{0.8}{
  \begin{tabular}{l|c|cc}
    \toprule
    Methods & \makecell[c]{Training w/\\EgoInt. Uncommon} & EgoInt. Uncommon & \makecell[c]{Zero-shot on\\RIO Uncommon} \\
    \midrule
    \multirow{2}{*}{Naive SFT} 
      & --         & 33.0 & 21.4 \\
      & \cellcolor[gray]{0.9}\checkmark & \cellcolor[gray]{0.9}\textbf{40.9} & \cellcolor[gray]{0.9}\textbf{22.1} \\
    \midrule
    \multirow{2}{*}{RoG SFT} 
      & --         & 33.7 & 23.8 \\
      & \cellcolor[gray]{0.9}\checkmark & \cellcolor[gray]{0.9}\textbf{44.7} & \cellcolor[gray]{0.9}\textbf{26.0} \\
    \bottomrule
  \end{tabular}
  }
  \label{tab:uncommon-intention-results}
\end{table}

\section{Conclusion}


We introduce egocentric visual intention grounding, where AI assistants infer and localize intended objects based on implicit human intentions. To support this research, we construct EgoIntention and benchmark state-of-the-art large vision-language models (LVLMs). Our results reveal that LVLMs struggle with implicit intention inference and egocentric visual grounding.
We propose Reason-to-Ground Instruction Tuning (RoG), a model-agnostic approach that disentangles intention reasoning from visual grounding, reducing spurious correlations and improving alignment with human intent.
By applying RoG in supervised fine-tuning with hybrid data from normal visual grounding and intention grounding tasks, LVLMs retain strong performance on conventional visual grounding while achieving significant improvements in egocentric visual intention grounding, offering a promising approach for both object and intention queries in exo- and ego-centric visual environments.
\section*{Acknowledgements}
Part of this work was carried out as a course project for CS6280: Deep Learning with Language Applications at the National University of Singapore. We thank Prof. Qizhe Xie for his helpful feedback during the course project.
{
    \small
    \bibliographystyle{ieeenat_fullname}
    \bibliography{main}
}


\end{document}